\documentclass[runningheads]{llncs}
\usepackage{graphicx}
\usepackage{subfig}
\usepackage{comment}
\usepackage{amsmath,amssymb} 
\usepackage{color}
\usepackage{algorithmicx}
\usepackage{algpseudocode}
\usepackage{algorithm}
\usepackage{booktabs}
\usepackage{hyperref}
\usepackage{bbm}
\usepackage{courier}
\usepackage{multirow}
\usepackage{textcomp}

\let\vec\boldsymbol

\begin{document}
\pagestyle{headings}
\mainmatter
\def\ECCVSubNumber{5772}  

\title{Meta-rPPG: Remote Heart Rate Estimation Using a Transductive Meta-Learner} 

\titlerunning{Remote Heart Rate Estimation Using a Transductive Meta-Learner}
%
\author{Eugene Lee \qquad
Evan Chen \qquad
Chen-Yi Lee}
\authorrunning{Lee et al.}
%
\institute{Institute of Electronics, National Chiao Tung University\\
\email{\{eugenelet.ee06g,evanchen.ee06\}@nctu.edu.tw, cylee@si2lab.org}}
\maketitle

\begin{abstract}
Remote heart rate estimation is the measurement of heart rate without any physical contact with the subject and is accomplished using remote photoplethysmography (rPPG) in this work. rPPG signals are usually collected using a video camera with a limitation of being sensitive to multiple contributing factors, e.g.\ variation in skin tone, lighting condition and facial structure. End-to-end supervised learning approach performs well when training data is abundant, covering a distribution that doesn't deviate too much from the distribution of testing data or during deployment. To cope with the unforeseeable distributional changes during deployment, we propose a transductive meta-learner that takes unlabeled samples during testing (deployment) for a self-supervised weight adjustment (also known as transductive inference), providing fast adaptation to the distributional changes. Using this approach, we achieve state-of-the-art performance on MAHNOB-HCI and UBFC-rPPG.

\end{abstract}

\section{Introduction}

Remote photoplethysmography (rPPG) is useful in situations where conventional approaches for heart rate estimation like electrocardiogram (ECG) and photoplethysmogram (PPG) that requires physical contact with the subject is infeasible. The acquisition of rPPG signal is useful for the estimation of physiological signal like heart rate (HR) and heart rate variation (HRV) which are important parameters for remote health-care. rPPG is usually obtained using a video camera and a growing number of studies have used rPPG for HR estimation \cite{poh2010non,poh2010advancements,balakrishnan2013detecting,de2013robust,li2014remote,tulyakov2016self,wang2016algorithmic}.

The adoption of deep learning techniques for the measurement of rPPG from face images is not novel and is supported by numerous studies \cite{chen2018deepphys,yu2019remote,niu2019rhythmnet,yu2019remote2}. All of the prior work involving deep learning uses an end-to-end supervised learning approach where a global model is deployed during inference (to the best of our knowledge), also known as inductive inference. It has been pointed out by the recent advances in deep learning that such approach doesn't perform well when there are changes in the modeled distribution when moving from training dataset to the real world \cite{bengio2019meta,finn2019online}. The unpredictability of the changes in environment and test subjects (skin-tone, facial structure, etc.) hinders the performance of remote heart rate estimation. To cope with the dynamical changes of the environment and subjects, we propose a transductive meta-learner that is able to perform fast adaption during deployment. Our algorithm introduces a warm-start time frame (2 seconds) for adaptation (weight update) to cope with the distributional changes, resulting in better performance during remote heart rate estimation.

Current advances in meta-learning \cite{finn2017model,nichol2018first,rusu2018meta} have shed light on the techniques of designing and training a deep neural network (DNN) that is able to adapt to new tasks through minimal weight changes. As prior work in meta-learning is built on well-defined tasks consisting of the classification of a few labeled examples (shots) provided as support set during test time for fast adaptation, it can't be directly applied to our context as labeled data are unobtainable during deployment. In our context, adaptation has to be done in a self-supervised fashion, hence we incorporate transductive inference techniques \cite{Hu2020Empirical,liu2018learning} into the design of our learning algorithm. The application of meta-learning and transductive inference to rPPG estimation is not trivial since we have to consider the modeling of both spatial and temporal information in the formulation of our meta-learner. In our work, we split our DNN into two parts: a feature extractor and a rPPG estimator modeled by a convolutional neural network (CNN) and long short-term memory (LSTM) \cite{hochreiter1997long} network respectively. We introduce a synthetic gradient generator modeled by a shallow Hourglass network \cite{newell2016stacked} along with a novel prototypical distance minimizer to perform transductive inference when labeled data are not available during deployment. Intuitively, the variation in distribution is caused by the visual aspect of the incoming signal, hence adaptation (weight changes) is only performed on the feature extractor while the rPPG estimator's parameters will be frozen during deployment. In summary, our main contributions are as follows:
\begin{enumerate}
    \item In Section \ref{sec:method}, we propose a meta-learning framework that exploits spatiotemporal information for remote heart rate estimation, specifically designed for fast adaptation to new video sequences during deployment.
    \item In Section \ref{sec:trans}, we introduce a transductive inference method that makes use of unlabeled data for fast adaptation during deployment using a \textit{synthetic gradient generator} and a novel \textit{prototypical distance minimizer}.
    \item In Section \ref{sec:ord}, we propose the formulation of rPPG estimation as an ordinal regression problem to cope with the mismatch in temporal domain of both visual input and collected PPG data, as studied in \cite{digiglio2014microflotronic}.
    \item In Section \ref{sec:exp}, we validate our proposed methods empirically on two public face heart rate dataset (MAHNOB-HCI \cite{soleymani2011multimodal} and UBFC-rPPG \cite{bobbia2019unsupervised}), trained using our collected dataset. Our experimental results demonstrate that our approach outperforms existing methods on remote heart rate estimation.
\end{enumerate}

\section{Related Work}\label{sec:related}
\subsubsection{Remote Heart Rate Estimation.}
The study of the measurement of heart rate using a video camera is not novel and has been supported by existing works. The remote measurement of heart rate is accomplished through the calculation of the cardiac pulse from the extracted rPPG signal which embeds the blood volume pulse (BVP). rPPG signal is extracted from the visible spectrum of light acquired by a video camera \cite{cennini2010heart}. Contact PPG sensors found in wearable devices and medical equipment estimates the heart rate by acquiring the light reflected from the skin using a light-emitting diode (LED) as its source. The constituents of the reflected light is composed of the amount of light absorbed by the skin (constant) and a time varying pulse signal contributed by the absorption of light by blood capillaries beneath the skin. Since existing contact PPG methods are non-intrusive, light sources with specific wavelengths are used, e.g.\ green light (525 nm) and infrared light (880 nm), having the properties of good absorption by blood and has minimal overlap with the visible light spectrum \cite{maeda2011advantages}. 

The feasibility of remote heart rate estimation was first proven by Verkruysse et al. \cite{verkruysse2008remote}, showing that PPG signals can be extracted from videos collected under an ambient light setting using webcams. To deal with noise, Poh et al.\ \cite{poh2010non} performed blind source separation on webcam filmed videos to extract pulse signal. A study is done in \cite{takano2007heart} revolving the comparison of the absorptivity of independent channels from the RGB channels by the human skin. It is concluded that the green channel is easily absorbed by the human skin, giving high signal-to-noise ratio for PPG signal acquisition. Based on the results in \cite{takano2007heart}, Li et al.\ \cite{li2014remote} used only the green channel for rPPG measurement and introduced an adaptive filtering technique (normalized least mean square) to eliminate motion artifact during deployment.

As it is hypothesized that all three channels (RGB) contain considerable information for the estimation of the rPPG signals, different channels are weighted and summed to receive better results when compared to using only a single channel (green) in \cite{mocco2016skin}. CHROM \cite{de2013robust} estimates the rPPG by using a method that linearly combines the RGB channels using the knowledge of a skin reflection model to separate pulse signal from motion-induced distortion. The POS \cite{wang2016algorithmic} method and the SB \cite{wang2017robust} model used the same skin reflection model as CHROM, but made a different assumption on the distortion model where another projection direction is used to separate the pulse and the distortions. All the mentioned methods are based on the calculation of the spatial mean of the entire face, assuming the contribution of each pixel to the estimation of rPPG is equal. Such assumption is not resilient to noise, rendering it infeasible in extreme conditions.

Recently, deep learning approaches have been introduced for rPPG signal estimation. DeepPhys \cite{chen2018deepphys} is an end-to-end supervised method implemented using a feed-forward CNN. Instead of using a Long Short-Term Memory (LSTM) cell to model the temporal information, an attention mechanism is used to learn the difference between frames. rPPGNet \cite{yu2019remote} considers the possibility of different types of video compression affecting the performance of rPPG estimation and proposed a network that handles both video quality reconstruction and rPPG signal estimation trained in an end-to-end fashion. RhythmNet \cite{niu2019rhythmnet} trained a CNN-RNN model based on a training set that contains diverse illumination and pose to enhance performance on public dataset. PhysNet \cite{yu2019remote2} constructed both a 3DCNN-based network and a RNN-based network and compared their performance.

\subsubsection{Meta-Learning.} The motivation for introducing meta-learning to our rPPG estimation framework is to perform fast adaptation of weights when our network is deployed in a setting that is not covered by our training distribution. Most of the studies revolving meta-learning are in few-shot classification, which have well-defined training and testing tasks. The structure of such well-defined tasks is exploited in \cite{mishra2017simple,santoro2016meta,vinyals2016matching} where the structure of the network is designed to take both training and test samples into consideration during inference. Gradient-based few-shot learning has also been proposed in \cite{finn2017model,nichol2018first,ravi2016optimization} with a limitation that the network capacity has to be small to prevent overfitting due to the small number of training samples. Few-shot learning based on the metric space has also been studied in \cite{koch2015siamese,vinyals2016matching,snell2017prototypical}. Meta-learning has also been applied to tasks beyond few-shot classification \cite{dou2019domain,yu2020foal,wu2019deep}, showing convincing results.

In our work, the parameters of our network is divided into two parts where one is responsible for fast adaptation and the other only responsible for estimation. Similar update methodology has been introduced in \cite{munkhdalai2017meta,gidaris2018dynamic,Hu2020Empirical,rusu2018meta,zintgraf2018fast}. All the proposed methods have different weight update orders and network construction but they have the consensus on the effectiveness of maintaining two sets of weights in a few-shot learning setting.

\section{Methodology}\label{sec:method}
\begin{figure}[!tbp]
\centering
  	\includegraphics[width=\textwidth]{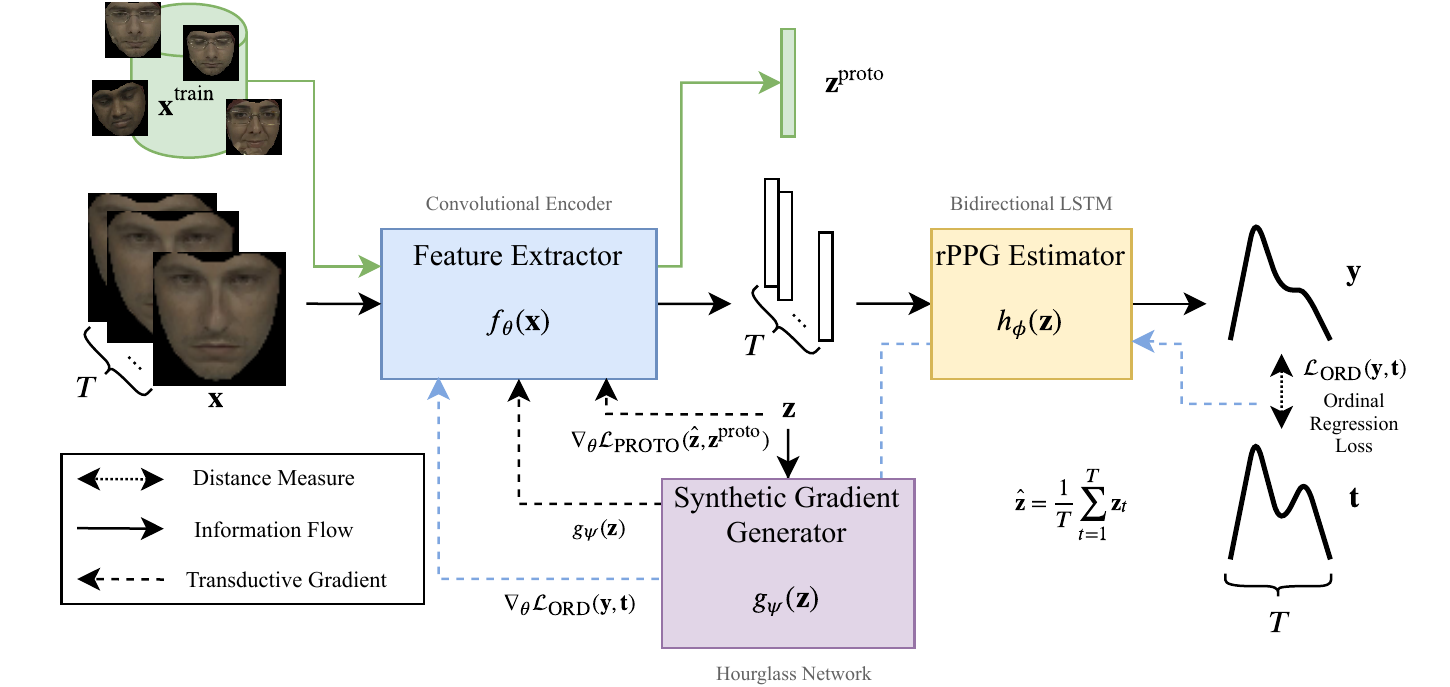}\label{fig:overview}
\caption{Overview of our system for transductive inference. Three modules: feature extractor, rPPG estimator and synthetic gradient generator are found in our system. During inference, only the feature extractor and rPPG estimator will be used. Only the parameters of the feature extractor, $\theta$, will be updated during transductive learning. Gradients used for the update of $\theta$ are shown in dashed lines where the blue dashed line is only present during the training phase of meta-learning. $\vec{z}^{\text{proto}}$ is generated using the training set, $\vec{x}^{\text{train}}$.}
\label{fig:overview}
\end{figure}

Given an input image from the camera, a face is first detected followed by the detection of facial landmarks. Pixel values within the landmarks are retained while the rest are filled with 0. The face image is then cropped and reshaped into a $K\times K$ image. We define the $i$-th video stream and PPG data containing a single subject as $\vec{x}^{(i)}$ and $\vec{t}^{(i)}$ respectively (we will drop the superscript $i$ whenever the context involves a single video stream for brevity) where $\vec{x}^{(i)}$ and $\vec{t}^{(i)}$ are sampled from a distribution of tasks $(\vec{x}^{(i)},\vec{t}^{(i)})\sim p(\mathcal{T})$. Continuous $T$ frames of face images are aggregated into a sequence, giving us the input data of our network, $\vec{x}_{t:t+T} \in \mathbb{R}^{K\times K \times T}$. The face sequences are paired with the PPG signal $\vec{t}_{t:t+T}\in \mathbb{R}^T$ collected using a PPG sensor placed beneath the index finger, where each sample $\vec{t}_t$ is temporally aligned to each frame of the face sequence $\vec{x}_t$ during the collection process. The output of our network is a rPPG signal $\vec{y}\in \mathbb{R}^T$ which is an estimation of the PPG signal having a small temporal offset caused by the carotid-radial pulse wave velocity (PWV) \cite{digiglio2014microflotronic}.


The training of our network is different from typical training practice found in an end-to-end supervised learning setting. We cast the learning of our network into a few-shot learning setting, involving both support set $\mathcal{S}$ and query set $\mathcal{Q}$. In a few-shot learning setting, $\mathcal{S}$ and $\mathcal{Q}$ are sampled from a large pool of tasks distribution $\{\mathcal{S}_n,\mathcal{Q}_n\} \sim \mathcal{T}$ where the sampled $\mathcal{S}_n$ and $\mathcal{Q}_n$ share the same set of classes but have disjoint samples for a classification problem. Our learning setting differs from the existing few-shot classification setting in two ways: 1.\ we are solving a regression problem, 2.\ our input is a sequence of images instead of still images that are distributed independent and identically.

In comparison to the typical few-shot learning setting, instead of sampling a set of samples originating from disjoint classes, we sample a sequence of image, $\vec{x}_{t:t+T}$, from disjoint video streams, e.g.\ videos containing different subject or background. During each sampling process, we sample $N$ independent video streams from our collected pool of video streams also known as distribution of tasks, $p(\mathcal{T})$. For each video stream, we split it into shorter sequences, where each sequence is further split into $V$ and $W$ frames. In correspondence to the few-shot learning setting, $V$ is the support set $\mathcal{S}$ used for adaptation and $W$ is the query set $\mathcal{Q}$ used to evaluate the performance of the model.

\subsection{Network Architecture}
Our meta-learning framework for remote heart rate estimation consists of three modules: convolutional encoder, rPPG estimator and synthetic gradient generator. To infer the rPPG signal, only the convolutional encoder and rPPG estimator will be used whereas the synthetic gradient generator will only be used during transductive learning. Our network is designed to exploit spatiotemporal information by first modeling visual information using a convolutional encoder followed by the modeling of the estimation of PPG signal using a LSTM rPPG estimator. We name our approach Meta-rPPG where detailed configuration of our architecture is shown in Table \ref{tab:arch}. An overview of our proposed framework is shown in Fig. \ref{fig:overview}.
\subsubsection{Convolutional Encoder.}
To extract latent features from a stream of images, we use a feature extractor modeled by a CNN, $f_\theta(\cdot)$. We use a ResNet-alike structure as the backbone of our convolutional encoder. Given an input stream of $T$ frames, the convolutional encoder is shared across frames:
\begin{equation}
    p_\theta(\vec{z}_i|\vec{x}_i)=f_\theta(\vec{x}_i),\label{eq:conv_dec}
\end{equation}
giving us $T$ independent distribution of latent features, $\{p_\theta(\vec{z}_i|\vec{x}_i)\}_{i=t}^{t+T}$, that will be fed to a rPPG estimator to model the temporal information for the estimation of rPPG signal.

\subsubsection{rPPG Estimator.} The latent features extracted from the input image stream are then passed to rPPG estimator modeled by a LSTM-MLP module, $h_\phi(\cdot)$. The intuition behind the split is to separate the parameters responsible for visual modeling from the parameters responsible for the estimation of rPPG signal, accomplished via the temporal modeling of the latent features:
\begin{equation}
    p_\phi(\vec{y}|\vec{z}_t,\vec{z}_{t+1},...,\vec{z}_{t+T})=h_\phi(\{p_\theta(\vec{z}_i|\vec{x}_i)\}_{i=t}^{t+T}).
\end{equation}
The LSTM module is designed to model the temporal information of a fixed sequence of $T$ latent features. The output of each step of the LSTM is followed by a MLP module which is responsible for the estimation of rPPG signal. As we model the estimation of rPPG signal as an ordinal regression task, we have a multitask ($S$ tasks) output. Details are deferred to Section \ref{sec:ord}.

\subsubsection{Synthetic Gradient Generator.}
For the fast adaptation of the parameters of our model during deployment, we introduce a synthetic gradient generator modeled by a shallow Hourglass network \cite{newell2016stacked}, $g_\psi(\cdot)$. The idea of using a synthetic gradient generator for transductive inference is not novel as it was first introduced in \cite{jaderberg2017decoupled} to parallelize the backpropagation of gradient and to augment backpropagation-through-time (BPTT) of long sequences found in LSTM. It is then applied to a few-shot learning framework in \cite{Hu2020Empirical} to generate gradient for unlabeled samples, giving a significant boost in performance. Our synthetic gradient generator attempts to model the gradient at $\vec{z}$ backpropagated from the ordinal regression loss from the output, $\mathcal{L}_{\text{ORD}}(\vec{y},\vec{t})$ (defined in (\ref{eq:ord_loss})):
\begin{equation}
    g_\psi (\vec{z}) = \nabla_{\vec{z}}\mathcal{L}_{\text{ORD}}(\vec{y},\vec{t}). \label{eq:syn_model}
\end{equation}
Our synthetic gradient generator will be used during transductive inference for the fast adaptation to new video sequences and at the adaptation phase during training, using the support set $\mathcal{S}$.

\begin{table}[]
\caption{Conv2DBlocks are composed of Conv2D, Batchnorm, average pooling, and ReLU. Conv1DBlocks are composed of Conv1D, Batchnorm and ReLU. Shortcut connections are added between Conv2DBlocks for the Convolutional Encoder. The synthetic gradient generator is designed as a Hourglass network. \checkmark\ indicates the information the layer acts upon. Output size is defined as $T\times$Channels$\times K\times K$ for Convolutional Encoder and $T\times$Channels for the rest.}
\centering
{\small
\begin{tabular}{ccccccc}
\toprule

Module & Layer & Output Size & Kernel Size & Spatial & Temporal\\ 

\midrule
\multirow{6}{*}{\parbox{2.3cm}{\centering Convolutional Encoder}}  
            &    Conv2DBlock & 60$\times$32$\times$32$\times$32 & 3$\times$3  & \checkmark \\
            &   Conv2DBlock & 60$\times$48$\times$16$\times$16 & 3$\times$3 & \checkmark\\
            &   Conv2DBlock & 60$\times$64$\times$8$\times$8 & 3$\times$3 & \checkmark\\
            &   Conv2DBlock & 60$\times$80$\times$4$\times$4 & 3$\times$3 & \checkmark\\
            &   Conv2DBlock & 60$\times$120$\times$2$\times$2 & 3$\times$3 & \checkmark\\
            &   AvgPool & 60$\times$120 & 2$\times$2 & \checkmark\\
\midrule
\multirow{3}{*}{\parbox{2.3cm}{\centering rPPG Estimator}}  
& Bidirectional LSTM & 60$\times$120 & - & \checkmark & \checkmark\\
&Linear & 60$\times$80 & - & \checkmark\\
&Ordinal & 60$\times$40 & - & \checkmark & \\
\midrule
\multirow{4}{*}{\parbox{2.3cm}{\centering Synthetic Gradient Generator}}  
&Conv1DBlock & 40$\times$120  & 3$\times$3 & \checkmark & \checkmark \\
&Conv1DBlock & 20$\times$120  & 3$\times$3 & \checkmark & \checkmark \\
&Conv1DBlock & 40$\times$120  & 3$\times$3 & \checkmark & \checkmark \\
&Conv1DBlock & 60$\times$120  & 3$\times$3&  \checkmark & \checkmark \\

 \bottomrule
\end{tabular}
}
\label{tab:arch}  
\end{table}

\subsection{Transductive Meta-Learning}\label{sec:trans}
During the deployment of Meta-rPPG, we have to consider the possibility of observing input samples that are not modeled by our model, also known as out-of-distribution samples. A possible solution is through the introduction of a pre-processing step that attempts to project the input data into a common distribution that is covered by our model. The consideration of an infinitely large distribution containing all possible scenarios during deployment is near impossible for any pre-processing technique. To cope with the shift in distribution, we propose two methods for transductive inference which provides gradient to our convolutional encoder $f_\theta(\cdot)$ during deployment. The first method is through the generation of synthetic gradients using a generator that is modeled during training. The second method attempts to minimize prototypical distance between different tasks which is based on the hypothesis that the rPPG estimator $h_\phi(\cdot)$ is only responsible for the estimation of rPPG signal. The modeling of the estimation of rPPG signal should not be affected by the visual input and is limited to a constrained distribution.

\subsubsection{Generating Synthetic Gradient.}
To generate gradients for weight update of our model during inference, we introduce a synthetic gradient generator that models the backpropagated gradient from the final output that uses an ordinal loss $\mathcal{L}_{\text{ORD}}(\vec{y},\vec{t})$ from (\ref{eq:ord_loss}), up to the output of the feature extractor $\vec{z}$, which can be simply put as $\nabla_{\vec{z}}\mathcal{L}_{\text{ORD}}(\vec{y},\vec{t})$. As our synthetic gradient generator $g_\psi (\vec{z})$ attempts to model $\nabla_{\vec{z}}\mathcal{L}_{\text{ORD}}(\vec{y},\vec{t})$, we can observe its role in the chain-rule of gradient update of the parameters of the feature extractor, $\theta$:
\begin{align}
    \theta &= \theta - \alpha \frac{\partial\mathcal{L}_{\text{ORD}}(\vec{y},\vec{t})}{\partial \vec{z}} \frac{\partial \vec{z}}{\partial \theta} \\
    &=  \theta - \alpha g_\psi (\mathbf{z})\frac{\partial \mathbf{z}}{\partial \theta}.
\end{align}
During the learning phase of training, we can update the weights of our synthetic gradient generator by minimizing the following objective function:
\begin{equation}
    \mathcal{L}_{\text{SYN}}(g_\psi(\vec{z}),\nabla_{\vec{z}}\mathcal{L}_{\text{ORD}}(\vec{y},\vec{t})) = ||g_\psi(\vec{z})-\nabla_{\vec{z}}\mathcal{L}_{\text{ORD}}(\vec{y},\vec{t})||_2^2, \label{eq:synthetic_loss}
\end{equation}
where the weight update of $\psi$ is given as:
\begin{equation}
    \psi = \psi - \eta \nabla_{\psi}\mathcal{L}_{\text{SYN}}(g_\psi(\vec{z}),\nabla_{\vec{z}}\mathcal{L}_{\text{ORD}}(\vec{y},\vec{t})).
\end{equation}

\subsubsection{Minimizing Prototypical Distance.}

\begin{figure}[!tbp]
\centering
     \subfloat[Prototypical distance minimization.\label{subfig-1:metric}]{%
       \includegraphics[width=0.45\textwidth]{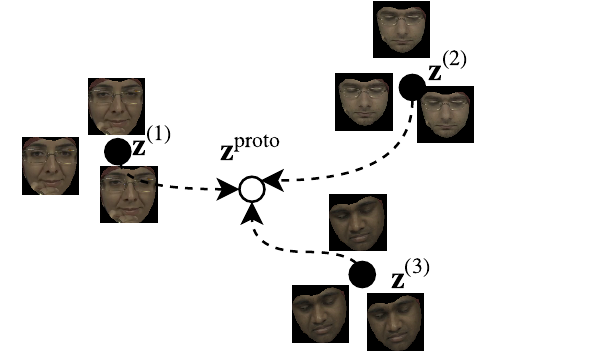}
     }
     \hfill
     \subfloat[Distribution of latent space.\label{subfig-2:distribution}]{%
       \includegraphics[width=0.45\textwidth]{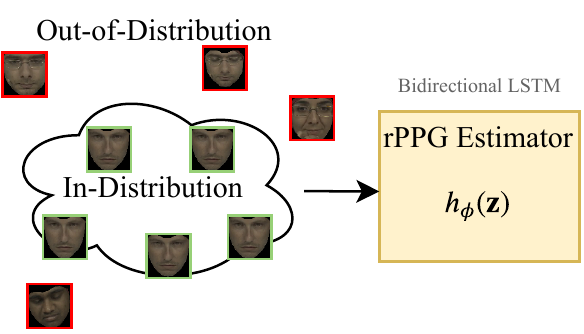}
     }
\caption{\protect\subref{subfig-1:metric} shows the high-level concept on the application of prototypical distance minimization on the latent space. \protect\subref{subfig-2:distribution} shows that the rPPG estimator only performs well on the in-distribution samples covered by the training dataset while performance on out-of-distribution samples is sub-optimal. In-distribution samples are outlined in green while out-of-distribution samples are outlined in red. The prototypical distance minimizer generates gradient that forces the out-of-distribution samples towards the center of the in-distribution samples.}
\label{fig:prototype_distance}
\end{figure}

As rPPG estimation is based on a visual input, there's no guarantee that the samples collected for training is consistent with the samples used during testing due to the broad distribution in the visual space. In a statistical viewpoint, data provided by the training set modeled by our network can be viewed as in-distribution samples whereas data collected under a different setting, e.g.\ lighting condition, subject and camera settings, are considered as out-of-distribution samples. It is not surprising that deep neural networks doesn't perform well on out-of-distribution samples as studied in \cite{liang2017enhancing,ren2019likelihood}. To overcome this limitation, we propose a prototypical distance minimization technique that can be applied on video sequences having a property where the statistical information that needs to be modeled by a neural network doesn't vary too much over time. We show a conceptual illustration in Fig. \ref{fig:prototype_distance}.

First, we consider each video sequences as separate task that needs to be modeled. We define the prototype of the latent variable of a specific task as:
\begin{equation}
    \vec{z}^{(i)} = \frac{1}{T}\sum_{t=1}^T p_\theta(\vec{z}_t^{(i)}|\vec{x}_t^{(i)}).
\end{equation}
Here, $T$ consecutive samples from task or video $i$ is sampled and the average is taken across the latent variable generated by the convolutional decoder from (\ref{eq:conv_dec}). We then obtain our first global latent variable prototype as:
\begin{align}
    \vec{z}^{\text{proto}} &= \mathbb{E}_{\vec{x}^{(i)}\sim p(\mathcal{T})}\frac{1}{T}\sum_{t=1}^T p_\theta(\vec{z}_t^{(i)}|\vec{x}_t^{(i)}) \\
    &= \mathbb{E}_{\vec{x}^{(i)}\sim p(\mathcal{T})}\frac{1}{T}\sum_{t=1}^T f_\theta(\vec{x}_t^{(i)}).
\end{align}
As mentioned earlier, we perform Monte Carlo sampling of $N$ tasks or videos for every training iteration, hence we are unable to obtain the statistical mean of the entire dataset in one shot. A more computationally feasible approach is to update our global latent variable prototype in an iterative fashion as:
\begin{equation}
    \vec{z}^{\text{proto}} = \gamma\vec{z}^{\text{proto}} + (1-\gamma)\mathbb{E}_{\vec{x}^{(i)}\sim p(\mathcal{T})}\frac{1}{T}\sum_{t=1}^T f_\theta(\vec{x}_t^{(i)}), \label{eq:proto_update}
\end{equation}
which can be understood as the weighted average between the old term and the newly sampled global latent variable prototype via the introduction of the hyperparameter $\gamma$. The global prototype in (\ref{eq:proto_update}) is obtained during the learning phase of training and transductive gradient is generated by minimizing the loss:
\begin{equation}
    \min_\theta\mathcal{L}_{\text{PROTO}}(\vec{z},\vec{z}^{\text{proto}}) = \min_\theta \mathbb{E}_{\vec{x}^{(i)}\sim p(\mathcal{T})} \frac{1}{T}\sum_{t=1}^T || p_\theta(\vec{z}_t^{(i)}|\vec{x}_t^{(i)}) - \vec{z}^{\text{proto}} ||_2^2. \label{eq:proto_loss}
\end{equation}

\subsubsection{Training of Meta-Learner.}
A meta-learner will perform well during testing if the setting during testing is similar to the setting during training. As mentioned earlier, we split the $N$ tasks samples from all our collected video into sequences that are further split into $V$ and $W$ frames. We use $V$ for the update of $\theta$ phrased as the \textit{adaptation phase} and $W$ for the update of $\phi$, $\psi$, $\theta$ and $\mathbf{z}^{\text{proto}}$ phrased as the \textit{learning phase}. In general, the split of $V$ and $W$ is put as $W>V$ and $W\cap V=\emptyset$ with the intuition that we attempt to minimize the frames required for adaptation and perform learning on the adapted space or distribution.

The role of the adaptation phase is to train our convolutional encoder $f_\theta(\cdot)$ to map input image sequences to a representation or latent space that will perform well when fed to our rPPG estimator $h_\phi(\cdot)$. During training, gradients from three sources will be used in the adaptation phase: synthetic gradient generator, prototypical distance minimizer and from the ordinal regression loss. During testing or deployment, we don't have any labeled data available, hence gradient from the ordinal regression loss is unattainable. Instead, we use gradients from our synthetic gradient generator and prototypical distance minimizer for adaptation. Note that the adaptation phase is run for $L$ steps on the same $V$ frames.

The role of the learning phase is to force our model to learn a suitable latent representation for rPPG signal estimation based on its image domain correspondence. Supervised learning is required in the learning phase, hence this phase will only be present during training. Here, $\theta$ and $\phi$ which corresponds to the convolutional encoder and rPPG estimator will be trained in an end-to-end fashion whereas the parameters of the synthetic gradient generator $\psi$ will be updated using the gradient backpropagated from the ordinal regression loss. $\psi$ is updated by minimizing the synthetic loss given in (\ref{eq:synthetic_loss}).

Before the inclusion of the adaptation phase during training, we first pre-train our network under the learning phase for $R$ epochs. The reason is that the gradient backpropagated to the synthetic gradient generator and the prototype used for the prototypical minimizer rely on the task at hand (rPPG estimation) rather than using gradient and prototype based on a set of random weights, which could lead to unstability. We summarize the training of our meta-learner in Algorithm \ref{alg:training}.

\subsection{Posing rPPG Estimation as an Ordinal Regression Task}\label{sec:ord}
Ordinal regression is commonly used in task that requires the prediction of labels that contains ordering information, e.g.\ age estimation \cite{niu2016ordinal,cao2019rank}, progression of various diseases \cite{doyle2014predicting,weersma2009molecular,streifler1995lack,sigrist2007progressive}, text message advertising \cite{rettie2005text} and various recommender systems \cite{parra2011implicit}. The motivation of using ordinal regression in our work is because there's an ordering of rPPG value that can be exploited and there will always be a temporal and magnitude discrepancy between rPPG and PPG signal as they originate from different parts of the human body \cite{digiglio2014microflotronic}.

To cast the estimation of rPPG signal as an ordinal regression problem, we first normalize a segment of PPG signal to be within 0 and 1. We then quantize it uniformly into 40 segments where each segment represents a \textit{rank}. Using the PPG segment $\vec{t}_{t:t+T}$ as an example, the quantized $t$-th sample will be categorized into one of the $S$ segments $\{\tau_1 \prec ... \prec \tau_S \}$. With a slight abuse of notations, if the categorized value for the $t$-th sample falls in the $s$-th segment, we denote it as $\vec{t}_{t,s}=\mathbbm{1}\{\vec{t}_t > \tau_s\}$ to keep our formulation concise. $\mathbbm{1}\{\cdot\}$ is an indicator function that returns 1 if the inner condition is true and 0 otherwise. As in \cite{niu2019rhythmnet}, our rPPG estimator will have to solve $S$ binary classification problem given as:
\begin{equation}
    \mathcal{L}_{\text{ORD}}(\vec{y},\vec{t}) =  - \frac{1}{T} \sum_{t=1}^T\sum_{s=1}^S \vec{t}_{t,s}\log (p_\phi(\vec{y}_{t,s}|\vec{z}_{t:t+T})) + (1-\vec{t}_{t,s})\log (1 - p_\phi(\vec{y}_{t,s}|\vec{z}_{t:t+T})). \label{eq:ord_loss}
\end{equation}
Note that in (\ref{eq:ord_loss}), the same notational abuse is applied to the output of our rPPG estimator, $\vec{y}$. During inference, we can obtain $T$ rPPG samples corresponding to $T$ consecutive frames, $\vec{x}_{t:t+T}$, fed to our model, giving us:
\begin{equation}
    \vec{y}_{t:t+T} = h_\phi(\vec{z}_{t:t+T})= \Bigl\{\sum_{s=1}^S \mathbbm{1}\{p_\phi (\vec{y}_{i,s}=1 | \vec{z}_{t:t+T})>0.5\}\Bigr\}_{i=t}^T.
\end{equation}

\begin{algorithm}[tb]
   \caption{Training of Meta-Learner}
   \label{alg:training}
\begin{algorithmic}[1]
   \State {\bfseries Input:} $p(\mathcal{T})$: distribution of tasks 
   \For{$i\gets 1, R$} \Comment{Pre-train network in an end-to-end fashion for $R$ epochs}
       \State Sample batch of tasks $\mathcal{T}_i\sim p(\mathcal{T})$ 
       \For{$(\vec{x},\vec{t})\sim\mathcal{T}_i$}
            \State Update $\theta$ and $\phi$ by minimizing $\mathcal{L}_{\text{ORD}} (\vec{y}, \vec{t})$ from (\ref{eq:ord_loss})
       \EndFor
   \EndFor
\While{not done} \Comment{Begin transductive meta-learning}
   \State Sample batch of tasks $\mathcal{T}_i\sim p(\mathcal{T})$ 
   \For{$(\vec{x},\vec{t})\sim\mathcal{T}_i$}
        \State $\{\hat{\vec{x}},\hat{\vec{t}}\}$, $\{\tilde{\vec{x}},\tilde{\vec{t}}\}$ $\leftarrow$ $\vec{x},\vec{t}$ \Comment{Split into $V$ and $W$ consecutive frames}
       \For{$i\gets 1, L$} \Comment{Adaptation phase (run $L$ steps)}
           \State $\theta \leftarrow \theta - \alpha (\nabla_{\theta}\mathcal{L}_{\text{proto}} (\hat{\vec{z}}, \hat{\vec{z}}^{\text{proto}}) + \nabla_\theta\mathcal{L}_{\text{ORD}} (\hat{\vec{y}}, \hat{\vec{t}})  + f_{\psi}(\hat{\vec{z}})) $
       \EndFor
       \State $\psi = \psi - \eta \nabla_{\psi}\mathcal{L}_{\text{SYN}}(f_\psi(\tilde{\vec{z}}),\nabla_{\tilde{\vec{z}}}\mathcal{L}_{\text{ORD}}(\tilde{\vec{y}},\tilde{\vec{t}}))$ \Comment{Learning phase}
       \State $\theta = \theta - \eta\nabla_\theta \mathcal{L}_{\text{ORD}}(\tilde{\vec{y}},\tilde{\vec{t}})$
       \State $\phi = \phi - \eta\nabla_\phi \mathcal{L}_{\text{ORD}}(\tilde{\vec{y}},\tilde{\vec{t}})$
       \State $\vec{z}^{\text{proto}} = \gamma\vec{z}^{\text{proto}} + (1-\gamma)\mathbb{E}_{\tilde{\vec{x}}^{(i)}\sim \tilde{\vec{x}}}\frac{1}{T}\sum_{t=1}^T f_\theta(\tilde{\vec{x}}_t^{(i)})$
   \EndFor
\EndWhile
\end{algorithmic}
\end{algorithm}

\section{Experiments}\label{sec:exp}
We show the efficacy of our proposed method by performing empirical simulation MAHNOB-HCI \cite{soleymani2011multimodal} and UBFC-rPPG \cite{bobbia2019unsupervised} using a model trained using our collected dataset. We also show how transductive inference helps in adapting to unseen datasets using a model trained using our own collected training dataset. Source code is available at \url{https://github.com/eugenelet/Meta-rPPG}.

\subsection{Dataset and Experimental Settings}
\subsubsection{MAHNOB-HCI dataset \cite{soleymani2011multimodal}} is recorded at 61 fps using a resolution of 780$\times$580 and includes 527 videos from 27 subjects. Since our training dataset is collected at 30 fps, we downsample the videos to 30 fps by getting rid half of the video frames. To generate ground truth heart rate (HR) for evaluation, we use the EXG1 signals containing ECG signal. Peaks of ECG signal are found using \texttt{scipy.signal.find\_peaks} and distance in samples between peaks is used for the calculation of HR. To compare fairly with previous works \cite{chen2018deepphys,niu2018synrhythm,vspetlik2018visual,yu2019remote2} [3, 12, 18], we follow the same routine in their works by using 30 seconds clip (frames 306 to 2135) of each video.

\subsubsection{UBFC-rPPG dataset \cite{bobbia2019unsupervised}} is relatively new and contains 38 uncompressed videos along with finger oximeter signals. Ground truth heart rate is provided, hence it can directly be used for evaluation without any additional processing. This dataset has a wider range of HR collected induced by making participants play a time-sensitive mathematical game that supposedly raises their HR. Diffuse reflections was created by introducing ambient light in the experiment.

\subsubsection{Collected Dataset.}
For the training of our model, we collected our own dataset. A RGB video camera is used for the collection of video at 30 fps using a resolution of 480$\times$640. PPG signals are collected using our self-designed device \cite{lee2019centralized} where raw data can be sent to another host device via a UART port. Collection of video and PPG signal are synced by connecting both devices to a Nvidia Jetson TX2. Since our PPG sensor device collects PPG signal at 100 Hz, we downsample it to 30 Hz to match the visual stream. Video and PPG samples of length over 2 minutes are collected and are cropped to 2 minutes to match our proposed meta-learning framework. A total of 19 videos are collected where 18 are used for training and 1 for validation.

\begin{figure}[!tbp]
\centering
  	\includegraphics[width=\textwidth]{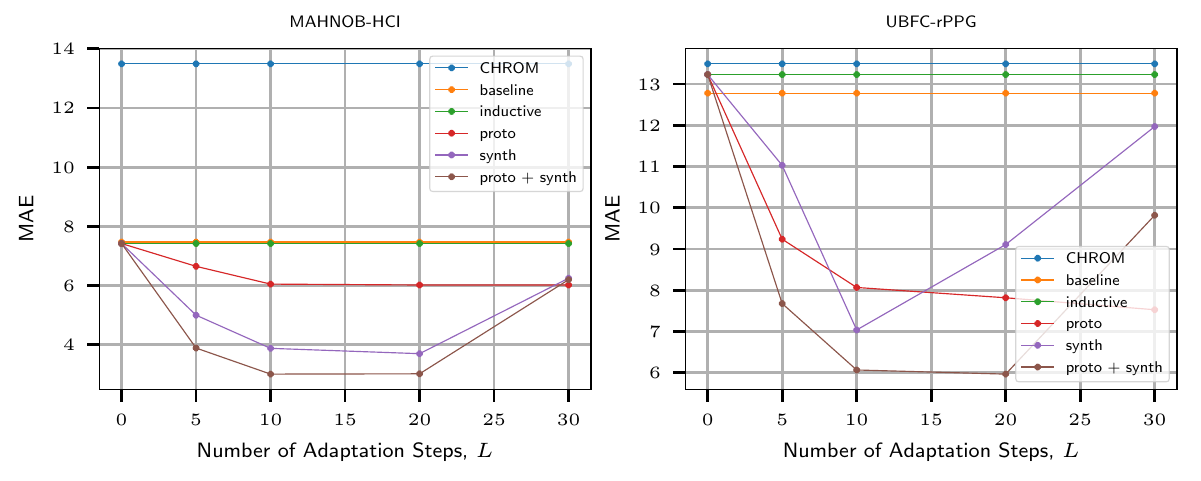}
\caption{MAE obtained using different rPPG estimation methods. Demonstrates how the number of adaptation steps, $L$, affects performance.}
\label{fig:performance}
\end{figure}

\subsubsection{Experimental Settings.}
Facial videos and corresponding PPG signals are synchronized before
training. For each video clip, we use the face detector found in \texttt{dlib} that uses \cite{dalal2005histograms} cascaded with a facial landmark detector implemented using \cite{kazemi2014one}. To keep the results of our face detector consistent, we use a median flow tracker provided by OpenCV \cite{opencv_library}. Pixels within the landmarks are retained and 5 pixels are used for zero-padding beyond the outermost pixels. The resulting face images are then resized to $64\times 64$. All experiments and network training are done on a single Nvidia GTX1080Ti using PyTorch \cite{NEURIPS2019_9015}. The SGD optimizer is used. We set the learning rate, $\eta=10^{-3}$, the adaptive learning rate, $\alpha=10^{-5}$ and the prototype update weight, $\gamma=0.8$. We train all the models for 20 epochs.

\subsubsection{Performance Metrics.} For the evaluation of public datasets, we report the error's standard deviation (SD), mean absolute error (MAE), root mean square error (RMSE) and Pearson's correlation coefficient (R).

\subsection{Evaluation on MAHNOB-HCI and UBFC-rPPG}
We evaluate 5 different configurations as part of the ablation study of the methods we introduced. The first configuration is the training of our proposed architecture in an end-to-end fashion and perform inductive inference on the test set, End-to-end (baseline). The second configuration is to train our network using all of our proposed methods but doesn't perform adaptation during inference, Meta-rPPG (inductive). The third and fourth configuration perform transductive inference be using either prototypical distance minimizer, Meta-rPPG (proto only), or synthetic gradient generator, Meta-rPPG (synth only), respectively. The final configuration uses both proposed method for transductive inference, Meta-rPPG (proto+synth).

Results of average HR measurement for MAHNOB-HCI and UBFC-rPPG are shown in Table \ref{tab:mahnob} and \ref{tab:ubfc} respectively.  We also show MAE results by the varying number of adaptation steps, $L$ in Fig. \ref{fig:performance}. Since we use $L=10$ during training, setting $L=10$ during evaluation also gives us the best results, which agrees with the common practice used in meta-learning \cite{finn2017model,nichol2018first}. Note that our proposed architecture used for end-to-end (baseline) training is similar to \cite{yu2019remote2} with a difference that we are using only 18 videos of length 2 minutes for training whereas \cite{yu2019remote2} uses OBF dataset \cite{li2018obf} that contains 212 videos of length 5 minutes. Considering the difference in magnitude of dataset size, it's understandable that training end-to-end using our network doesn't perform as well as in \cite{yu2019remote2}. This also indicates that our performance can be further improved just by collecting more data. More experimental results are shown in the Supplementary Materials.

\begin{table}[]
\caption{Results of average HR measurement on MAHNOB-HCI.}
\centering
{\scriptsize
\begin{tabular}{lcccc}
\toprule
\multirow{2}{*}{Method} & \multicolumn{4}{c}{HR (bpm)} \\
                                        & SD        & MAE       & RMSE      & R \\
\midrule
Poh2011 \cite{poh2010advancements}      &   13.5    &   -       &   13.6    &   0.36 \\
CHROM \cite{de2013robust}               &   -       &   13.49   &   22.36   &   0.21 \\
Li2014 \cite{li2014remote}              &   6.88    &   -       &   7.62    &   0.81  \\
SAMC \cite{tulyakov2016self}            &   5.81    &   -       &   6.23    &   0.83   \\
SynRhythm \cite{niu2018synrhythm}       &   10.88   &   -       &   11.08   &   -   \\
HR-CNN \cite{vspetlik2018visual}        &   -       &   7.25    &   9.24    &   0.51    \\
DeepPhys \cite{chen2018deepphys}        &   -       &   4.57    &   -       &   -   \\
PhysNet \cite{yu2019remote2}            &   7.84    &   5.96    &   7.88    &   0.76  \\
STVEN+rPPGNet \cite{yu2019remote}       &   5.57    &   4.03    &   5.93    &   \textbf{0.88} \\
\midrule
End-to-end (baseline)                   &   7.39    &   7.47    &   8.63    &   0.70   \\
Meta-rPPG (inductive)                   &   7.91    &   7.42    &   8.65    &   0.74   \\
Meta-rPPG (proto only)                       &   6.89    &   6.05    &   6.71    &   0.77    \\
Meta-rPPG (synth only)                       &   5.09    &   3.88    &   4.02    &   0.81   \\
Meta-rPPG (proto+synth)               &   \textbf{4.90}    &   \textbf{3.01}    &   \textbf{3.68}    &   0.85   \\

 \bottomrule
\end{tabular}
}
\label{tab:mahnob}  
\end{table}

\begin{table}[]
\caption{Results of average HR measurement on UBFC-rPPG.}
\centering
{\scriptsize
\begin{tabular}{lcccc}
\toprule
\multirow{2}{*}{Method} & \multicolumn{4}{c}{HR (bpm)} \\
                                        &   SD      &   MAE         & RMSE      & R     \\ 
\midrule
GREEN \cite{verkruysse2008remote}       &   20.2    &   10.2       &   20.6    &   -    \\
ICA \cite{poh2010advancements}          &   18.6    &   8.43       &   18.8    &   -    \\
CHROM \cite{de2013robust}               &   19.1    &   10.6        &   20.3   &   -    \\
POS \cite{wang2016algorithmic}          &   10.4    &   \textbf{4.12}       &   10.5    & -\\
3D CNN \cite{bousefsaf20193d}           &   8.55    &   5.45       &   8.64   &   -   \\
\midrule
End-to-end (baseline)                   &   13.70    &   12.78    &   13.30    &   0.27   \\
Meta-rPPG (inductive)                   &   14.17    &   13.23    &   14.63    &   0.35   \\
Meta-rPPG (proto only)                       &   9.17    &   7.82    &   9.37    &   0.48   \\
Meta-rPPG (synth only)                       &   11.92    &   9.11    &   11.55    &   0.42   \\
Meta-rPPG (proto+synth)               &   \textbf{7.12}    &   5.97    &   \textbf{7.42}    &   \textbf{0.53}\\

 \bottomrule
\end{tabular}
}
\label{tab:ubfc}  
\end{table}

\subsubsection{Visualization of feature activation map}
shows the importance of transductive inference during evaluation. In Fig. \ref{fig:fvm}, we show feature activation maps visualization \cite{menikdiwela2017cnn} generated using our proposed methods and supervised end-to-end trained model on images extracted from MAHNOB-HCI. Regions that are more likely to be useful for remote HR estimation are more pronounced when our approach is introduced. Results show that transductive inference is useful when applied to data excluded from the training distribution.

\begin{figure}[!tbp]
\centering
  	\includegraphics[width=\textwidth]{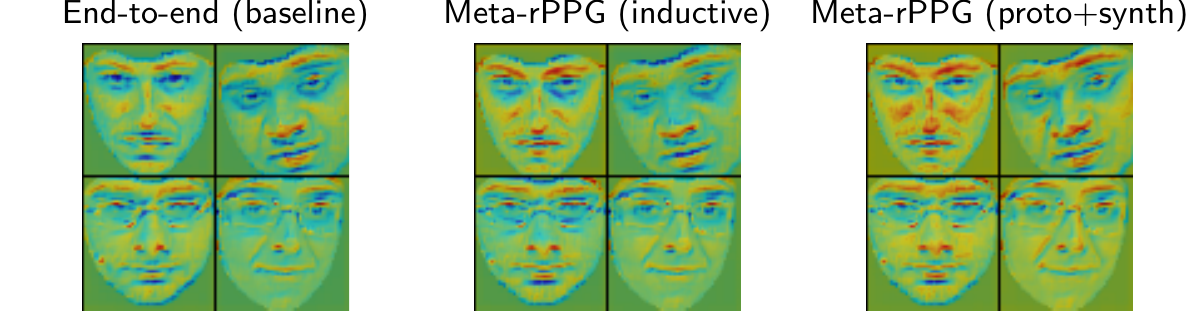}
\caption{Feature activation map visualization of 4 subjects using different training methods. Usage of transductive inference results in activations of higher contrast and covers larger region of facial features that contributes to rPPG estimation.}
\label{fig:fvm}
\end{figure}

\section{Conclusion}\label{sec:conc}
In this work, we introduce transductive inference into the framework of rPPG estimation. For transductive inference, we propose the use of a synthetic gradient generator and protypical distance minimizer to provide gradient to our feature extractor when labeled data are unobtainable. By posing the learning of our network as a meta-learning framework, we see substantial improvements on MAHNOB-HCI and UBFC-rPPG dataset demonstrating state-of-the-art results.

\subsubsection*{Acknowledgements.} This work is supported by Ministry of Science and Technology (MOST) of Taiwan: 107-2221-E-009 -125 -MY3.

\clearpage
%
%
\bibliographystyle{splncs04}
\bibliography{egbib}

\begin{thebibliography}{10}
\providecommand{\url}[1]{\texttt{#1}}
\providecommand{\urlprefix}{URL }
\providecommand{\doi}[1]{https://doi.org/#1}

\bibitem{balakrishnan2013detecting}
Balakrishnan, G., Durand, F., Guttag, J.: Detecting pulse from head motions in
  video. In: Proceedings of the IEEE Conference on Computer Vision and Pattern
  Recognition. pp. 3430--3437 (2013)

\bibitem{bengio2019meta}
Bengio, Y., Deleu, T., Rahaman, N., Ke, R., Lachapelle, S., Bilaniuk, O.,
  Goyal, A., Pal, C.: A meta-transfer objective for learning to disentangle
  causal mechanisms. arXiv preprint arXiv:1901.10912  (2019)

\bibitem{bobbia2019unsupervised}
Bobbia, S., Macwan, R., Benezeth, Y., Mansouri, A., Dubois, J.: Unsupervised
  skin tissue segmentation for remote photoplethysmography. Pattern Recognition
  Letters  \textbf{124},  82--90 (2019)

\bibitem{bousefsaf20193d}
Bousefsaf, F., Pruski, A., Maaoui, C.: 3d convolutional neural networks for
  remote pulse rate measurement and mapping from facial video. Applied Sciences
   \textbf{9}(20), ~4364 (2019)

\bibitem{opencv_library}
Bradski, G.: {The OpenCV Library}. Dr. Dobb's Journal of Software Tools  (2000)

\bibitem{cao2019rank}
Cao, W., Mirjalili, V., Raschka, S.: Rank-consistent ordinal regression for
  neural networks. arXiv preprint arXiv:1901.07884  (2019)

\bibitem{cennini2010heart}
Cennini, G., Arguel, J., Ak{\c{s}}it, K., van Leest, A.: Heart rate monitoring
  via remote photoplethysmography with motion artifacts reduction. Optics
  express  \textbf{18}(5),  4867--4875 (2010)

\bibitem{chen2018deepphys}
Chen, W., McDuff, D.: Deepphys: Video-based physiological measurement using
  convolutional attention networks. In: Proceedings of the European Conference
  on Computer Vision (ECCV). pp. 349--365 (2018)

\bibitem{dalal2005histograms}
Dalal, N., Triggs, B.: Histograms of oriented gradients for human detection.
  In: 2005 IEEE computer society conference on computer vision and pattern
  recognition (CVPR'05). vol.~1, pp. 886--893. IEEE (2005)

\bibitem{de2013robust}
De~Haan, G., Jeanne, V.: Robust pulse rate from chrominance-based rppg. IEEE
  Transactions on Biomedical Engineering  \textbf{60}(10),  2878--2886 (2013)

\bibitem{digiglio2014microflotronic}
Digiglio, P., Li, R., Wang, W., Pan, T.: Microflotronic arterial tonometry for
  continuous wearable non-invasive hemodynamic monitoring. Annals of biomedical
  engineering  \textbf{42}(11),  2278--2288 (2014)

\bibitem{dou2019domain}
Dou, Q., de~Castro, D.C., Kamnitsas, K., Glocker, B.: Domain generalization via
  model-agnostic learning of semantic features. In: Advances in Neural
  Information Processing Systems. pp. 6450--6461 (2019)

\bibitem{doyle2014predicting}
Doyle, O.M., Westman, E., Marquand, A.F., Mecocci, P., Vellas, B., Tsolaki, M.,
  K{\l}oszewska, I., Soininen, H., Lovestone, S., Williams, S.C., et~al.:
  Predicting progression of alzheimer’s disease using ordinal regression.
  PloS one  \textbf{9}(8) (2014)

\bibitem{finn2017model}
Finn, C., Abbeel, P., Levine, S.: Model-agnostic meta-learning for fast
  adaptation of deep networks. In: Proceedings of the 34th International
  Conference on Machine Learning-Volume 70. pp. 1126--1135. JMLR. org (2017)

\bibitem{finn2019online}
Finn, C., Rajeswaran, A., Kakade, S., Levine, S.: Online meta-learning. arXiv
  preprint arXiv:1902.08438  (2019)

\bibitem{gidaris2018dynamic}
Gidaris, S., Komodakis, N.: Dynamic few-shot visual learning without
  forgetting. In: Proceedings of the IEEE Conference on Computer Vision and
  Pattern Recognition. pp. 4367--4375 (2018)

\bibitem{hochreiter1997long}
Hochreiter, S., Schmidhuber, J.: Long short-term memory. Neural computation
  \textbf{9}(8),  1735--1780 (1997)

\bibitem{Hu2020Empirical}
Hu, S.X., Moreno, P., Xiao, Y., Shen, X., Obozinski, G., Lawrence, N.,
  Damianou, A.: Empirical bayes transductive meta-learning with synthetic
  gradients. In: International Conference on Learning Representations (ICLR)
  (2020), \url{https://openreview.net/forum?id=Hkg-xgrYvH}

\bibitem{jaderberg2017decoupled}
Jaderberg, M., Czarnecki, W.M., Osindero, S., Vinyals, O., Graves, A., Silver,
  D., Kavukcuoglu, K.: Decoupled neural interfaces using synthetic gradients.
  In: Proceedings of the 34th International Conference on Machine
  Learning-Volume 70. pp. 1627--1635. JMLR. org (2017)

\bibitem{kazemi2014one}
Kazemi, V., Sullivan, J.: One millisecond face alignment with an ensemble of
  regression trees. In: Proceedings of the IEEE conference on computer vision
  and pattern recognition. pp. 1867--1874 (2014)

\bibitem{koch2015siamese}
Koch, G., Zemel, R., Salakhutdinov, R.: Siamese neural networks for one-shot
  image recognition. In: ICML deep learning workshop. vol.~2. Lille (2015)

\bibitem{lee2019centralized}
Lee, E., Hsu, T.J., Lee, C.Y.: Centralized state sensing using sensor array on
  wearable device. In: 2019 IEEE International Symposium on Circuits and
  Systems (ISCAS). pp.~1--5. IEEE (2019)

\bibitem{li2018obf}
Li, X., Alikhani, I., Shi, J., Seppanen, T., Junttila, J., Majamaa-Voltti, K.,
  Tulppo, M., Zhao, G.: The obf database: A large face video database for
  remote physiological signal measurement and atrial fibrillation detection.
  In: 2018 13th IEEE International Conference on Automatic Face \& Gesture
  Recognition (FG 2018). pp. 242--249. IEEE (2018)

\bibitem{li2014remote}
Li, X., Chen, J., Zhao, G., Pietikainen, M.: Remote heart rate measurement from
  face videos under realistic situations. In: Proceedings of the IEEE
  conference on computer vision and pattern recognition. pp. 4264--4271 (2014)

\bibitem{liang2017enhancing}
Liang, S., Li, Y., Srikant, R.: Enhancing the reliability of
  out-of-distribution image detection in neural networks. arXiv preprint
  arXiv:1706.02690  (2017)

\bibitem{liu2018learning}
Liu, Y., Lee, J., Park, M., Kim, S., Yang, E., Hwang, S.J., Yang, Y.: Learning
  to propagate labels: Transductive propagation network for few-shot learning.
  arXiv preprint arXiv:1805.10002  (2018)

\bibitem{maeda2011advantages}
Maeda, Y., Sekine, M., Tamura, T.: The advantages of wearable green reflected
  photoplethysmography. Journal of medical systems  \textbf{35}(5),  829--834
  (2011)

\bibitem{menikdiwela2017cnn}
Menikdiwela, M., Nguyen, C., Li, H., Shaw, M.: Cnn-based small object detection
  and visualization with feature activation mapping. In: 2017 International
  Conference on Image and Vision Computing New Zealand (IVCNZ). pp.~1--5. IEEE
  (2017)

\bibitem{mishra2017simple}
Mishra, N., Rohaninejad, M., Chen, X., Abbeel, P.: A simple neural attentive
  meta-learner. arXiv preprint arXiv:1707.03141  (2017)

\bibitem{mocco2016skin}
Mo{\c{c}}o, A.V., Stuijk, S., de~Haan, G.: Skin inhomogeneity as a source of
  error in remote ppg-imaging. Biomedical optics express  \textbf{7}(11),
  4718--4733 (2016)

\bibitem{munkhdalai2017meta}
Munkhdalai, T., Yu, H.: Meta networks. In: Proceedings of the 34th
  International Conference on Machine Learning-Volume 70. pp. 2554--2563. JMLR.
  org (2017)

\bibitem{newell2016stacked}
Newell, A., Yang, K., Deng, J.: Stacked hourglass networks for human pose
  estimation. In: European conference on computer vision. pp. 483--499.
  Springer (2016)

\bibitem{nichol2018first}
Nichol, A., Achiam, J., Schulman, J.: On first-order meta-learning algorithms.
  arXiv preprint arXiv:1803.02999  (2018)

\bibitem{niu2018synrhythm}
Niu, X., Han, H., Shan, S., Chen, X.: Synrhythm: Learning a deep heart rate
  estimator from general to specific. In: 2018 24th International Conference on
  Pattern Recognition (ICPR). pp. 3580--3585. IEEE (2018)

\bibitem{niu2019rhythmnet}
Niu, X., Shan, S., Han, H., Chen, X.: Rhythmnet: End-to-end heart rate
  estimation from face via spatial-temporal representation. IEEE Transactions
  on Image Processing  (2019)

\bibitem{niu2016ordinal}
Niu, Z., Zhou, M., Wang, L., Gao, X., Hua, G.: Ordinal regression with multiple
  output cnn for age estimation. In: Proceedings of the IEEE conference on
  computer vision and pattern recognition. pp. 4920--4928 (2016)

\bibitem{parra2011implicit}
Parra, D., Karatzoglou, A., Amatriain, X., Yavuz, I.: Implicit feedback
  recommendation via implicit-to-explicit ordinal logistic regression mapping.
  Proceedings of the CARS-2011  \textbf{5} (2011)

\bibitem{NEURIPS2019_9015}
Paszke, A., Gross, S., Massa, F., Lerer, A., Bradbury, J., Chanan, G., Killeen,
  T., Lin, Z., Gimelshein, N., Antiga, L., Desmaison, A., Kopf, A., Yang, E.,
  DeVito, Z., Raison, M., Tejani, A., Chilamkurthy, S., Steiner, B., Fang, L.,
  Bai, J., Chintala, S.: Pytorch: An imperative style, high-performance deep
  learning library. In: Wallach, H., Larochelle, H., Beygelzimer, A.,
  d\textquotesingle Alch\'{e}-Buc, F., Fox, E., Garnett, R. (eds.) Advances in
  Neural Information Processing Systems 32, pp. 8024--8035. Curran Associates,
  Inc. (2019),
  \url{http://papers.neurips.cc/paper/9015-pytorch-an-imperative-style-high-performance-deep-learning-library.pdf}

\bibitem{poh2010advancements}
Poh, M.Z., McDuff, D.J., Picard, R.W.: Advancements in noncontact,
  multiparameter physiological measurements using a webcam. IEEE transactions
  on biomedical engineering  \textbf{58}(1),  7--11 (2010)

\bibitem{poh2010non}
Poh, M.Z., McDuff, D.J., Picard, R.W.: Non-contact, automated cardiac pulse
  measurements using video imaging and blind source separation. Optics express
  \textbf{18}(10),  10762--10774 (2010)

\bibitem{ravi2016optimization}
Ravi, S., Larochelle, H.: Optimization as a model for few-shot learning  (2016)

\bibitem{ren2019likelihood}
Ren, J., Liu, P.J., Fertig, E., Snoek, J., Poplin, R., Depristo, M., Dillon,
  J., Lakshminarayanan, B.: Likelihood ratios for out-of-distribution
  detection. In: Advances in Neural Information Processing Systems. pp.
  14680--14691 (2019)

\bibitem{rettie2005text}
Rettie, R., Grandcolas, U., Deakins, B.: Text message advertising: Response
  rates and branding effects. Journal of targeting, measurement and analysis
  for marketing  \textbf{13}(4),  304--312 (2005)

\bibitem{rusu2018meta}
Rusu, A.A., Rao, D., Sygnowski, J., Vinyals, O., Pascanu, R., Osindero, S.,
  Hadsell, R.: Meta-learning with latent embedding optimization. arXiv preprint
  arXiv:1807.05960  (2018)

\bibitem{santoro2016meta}
Santoro, A., Bartunov, S., Botvinick, M., Wierstra, D., Lillicrap, T.:
  Meta-learning with memory-augmented neural networks. In: International
  conference on machine learning. pp. 1842--1850 (2016)

\bibitem{sigrist2007progressive}
Sigrist, M.K., Taal, M.W., Bungay, P., McIntyre, C.W.: Progressive vascular
  calcification over 2 years is associated with arterial stiffening and
  increased mortality in patients with stages 4 and 5 chronic kidney disease.
  Clinical Journal of the American Society of Nephrology  \textbf{2}(6),
  1241--1248 (2007)

\bibitem{snell2017prototypical}
Snell, J., Swersky, K., Zemel, R.: Prototypical networks for few-shot learning.
  In: Advances in neural information processing systems. pp. 4077--4087 (2017)

\bibitem{soleymani2011multimodal}
Soleymani, M., Lichtenauer, J., Pun, T., Pantic, M.: A multimodal database for
  affect recognition and implicit tagging. IEEE transactions on affective
  computing  \textbf{3}(1),  42--55 (2011)

\bibitem{vspetlik2018visual}
{\v{S}}petl{\'\i}k, R., Franc, V., Matas, J.: Visual heart rate estimation with
  convolutional neural network. In: Proceedings of the British Machine Vision
  Conference, Newcastle, UK. pp.~3--6 (2018)

\bibitem{streifler1995lack}
Streifler, J.Y., Eliasziw, M., Benavente, O.R., Hachinski, V.C., Fox, A.J.,
  Barnett, H.: Lack of relationship between leukoaraiosis and carotid artery
  disease. Archives of neurology  \textbf{52}(1),  21--24 (1995)

\bibitem{takano2007heart}
Takano, C., Ohta, Y.: Heart rate measurement based on a time-lapse image.
  Medical engineering \& physics  \textbf{29}(8),  853--857 (2007)

\bibitem{tulyakov2016self}
Tulyakov, S., Alameda-Pineda, X., Ricci, E., Yin, L., Cohn, J.F., Sebe, N.:
  Self-adaptive matrix completion for heart rate estimation from face videos
  under realistic conditions. In: Proceedings of the IEEE Conference on
  Computer Vision and Pattern Recognition. pp. 2396--2404 (2016)

\bibitem{verkruysse2008remote}
Verkruysse, W., Svaasand, L.O., Nelson, J.S.: Remote plethysmographic imaging
  using ambient light. Optics express  \textbf{16}(26),  21434--21445 (2008)

\bibitem{vinyals2016matching}
Vinyals, O., Blundell, C., Lillicrap, T., Wierstra, D., et~al.: Matching
  networks for one shot learning. In: Advances in neural information processing
  systems. pp. 3630--3638 (2016)

\bibitem{wang2016algorithmic}
Wang, W., den Brinker, A.C., Stuijk, S., de~Haan, G.: Algorithmic principles of
  remote ppg. IEEE Transactions on Biomedical Engineering  \textbf{64}(7),
  1479--1491 (2016)

\bibitem{wang2017robust}
Wang, W., den Brinker, A.C., Stuijk, S., de~Haan, G.: Robust heart rate from
  fitness videos. Physiological measurement  \textbf{38}(6), ~1023 (2017)

\bibitem{weersma2009molecular}
Weersma, R.K., Stokkers, P.C., van Bodegraven, A.A., van Hogezand, R.A.,
  Verspaget, H.W., de~Jong, D.J., Van Der~Woude, C., Oldenburg, B., Linskens,
  R., Festen, E., et~al.: Molecular prediction of disease risk and severity in
  a large dutch crohn’s disease cohort. Gut  \textbf{58}(3),  388--395 (2009)

\bibitem{wu2019deep}
Wu, Y., Rosca, M., Lillicrap, T.: Deep compressed sensing. arXiv preprint
  arXiv:1905.06723  (2019)

\bibitem{yu2020foal}
Yu, H., Sun, S., Yu, H., Chen, X., Shi, H., Huang, T.S., Chen, T.: Foal: Fast
  online adaptive learning for cardiac motion estimation. In: Proceedings of
  the IEEE/CVF Conference on Computer Vision and Pattern Recognition. pp.
  4313--4323 (2020)

\bibitem{yu2019remote2}
Yu, Z., Li, X., Zhao, G.: Remote photoplethysmograph signal measurement from
  facial videos using spatio-temporal networks. In: Proc. BMVC. pp. 1--12
  (2019)

\bibitem{yu2019remote}
Yu, Z., Peng, W., Li, X., Hong, X., Zhao, G.: Remote heart rate measurement
  from highly compressed facial videos: an end-to-end deep learning solution
  with video enhancement. In: Proceedings of the IEEE International Conference
  on Computer Vision. pp. 151--160 (2019)

\bibitem{zintgraf2018fast}
Zintgraf, L.M., Shiarlis, K., Kurin, V., Hofmann, K., Whiteson, S.: Fast
  context adaptation via meta-learning. arXiv preprint arXiv:1810.03642  (2018)

\end{thebibliography}

\clearpage

\appendix

\section{Performing Transductive Inference During Deployment}
Here, we show the algorithm for transductive inference during deployment in Algorithm \ref{alg:deploy}. The inference process is similar to the typical inference of a feed-forward deep neural network (the final line). The difference is in the inclusion of adaptation steps using the first 2 seconds of the video for transductive learning prior to actual inference.

\begin{algorithm}[tb]
   \caption{Transductive Inference During Deployment}
   \label{alg:deploy}
\begin{algorithmic}[1]
   \State {\bfseries Input:} $\vec{x}$: A single video stream
    \State $\hat{\vec{x}}$, $\tilde{\vec{x}}$ $\leftarrow$ $\vec{x}$ \Comment{$\hat{\vec{x}}$: first 2 seconds of video, $\tilde{\vec{x}}$: rest of video}
   \For{$i\gets 1, L$} \Comment{Adaptation phase (run $L$ steps)}
       \State $\theta \leftarrow \theta - \alpha (\nabla_{\theta}\mathcal{L}_{\text{proto}} (\hat{\vec{z}}, \hat{\vec{z}}^{\text{proto}})  + f_{\psi}(\hat{\vec{z}})) $
   \EndFor
   \State $\vec{y} \leftarrow h_{\phi}(f_\theta(\tilde{\vec{x}})) $ \Comment{Estimation of rPPG signal using adapted feature extractor}
\end{algorithmic}
\end{algorithm}

\section{Performance Comparison Using Different Adaptation Steps}
\begin{figure}[!tbp]
\centering
  	\includegraphics[width=\textwidth]{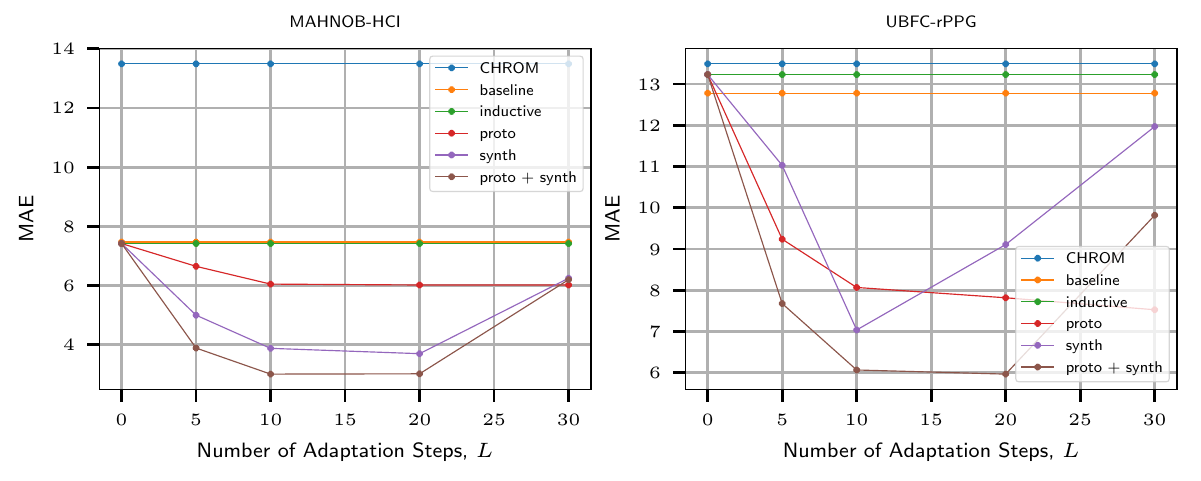}
\caption{Mean absolute error, MAE, obtained using different rPPG estimation methods. Demonstrates how the number of adaptation steps, $L$, affects performance.}
\label{fig:mae}
\end{figure}

\begin{figure}[!tbp]
\centering
  	\includegraphics[width=\textwidth]{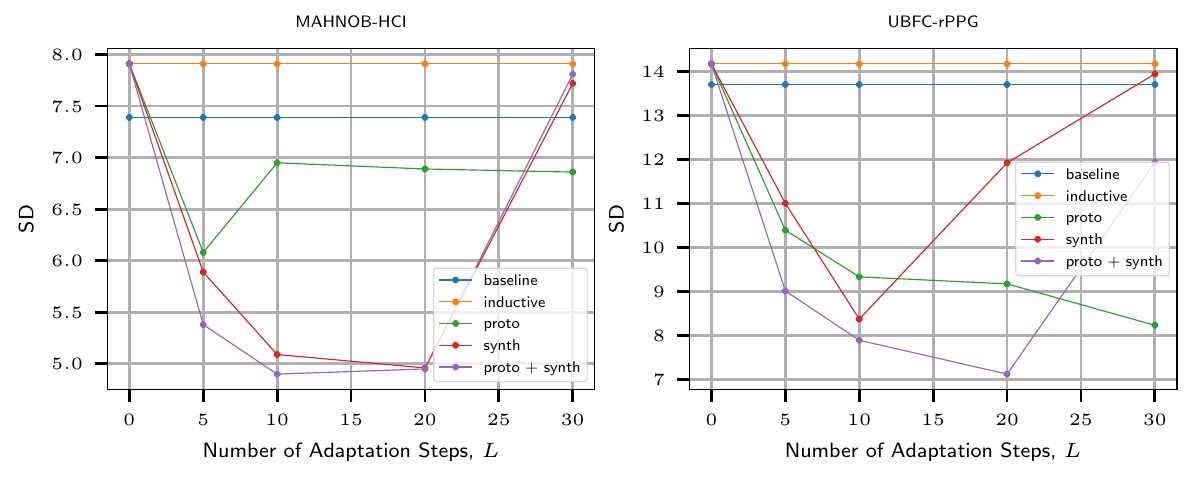}
\caption{Standard deviation of error, SD, obtained using different rPPG estimation methods. Demonstrates how the number of adaptation steps, $L$, affects performance.}
\label{fig:sd}
\end{figure}

\begin{figure}[!tbp]
\centering
  	\includegraphics[width=\textwidth]{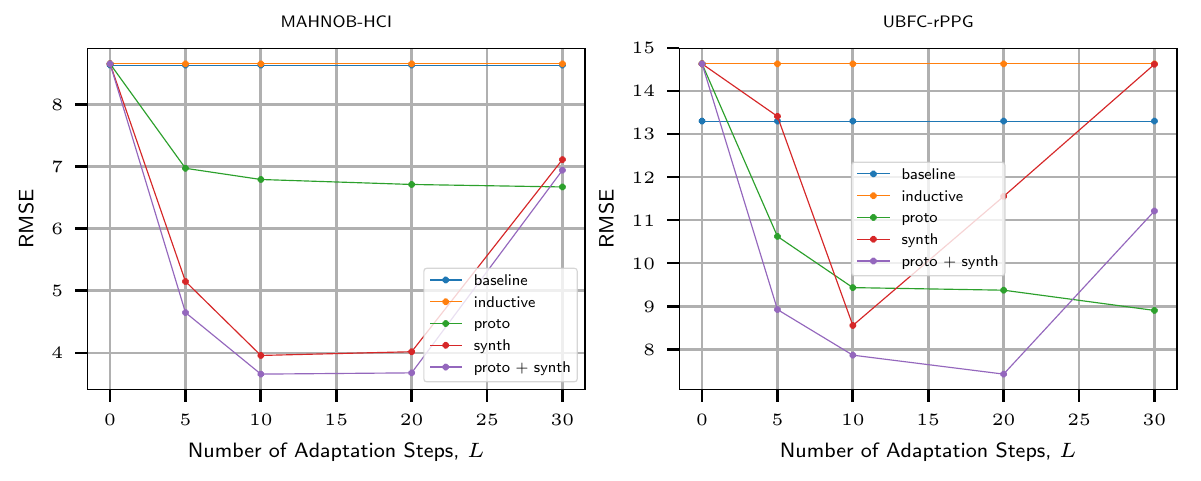}
\caption{Root mean squared error, RMSE, obtained using different rPPG estimation methods. Demonstrates how the number of adaptation steps, $L$, affects performance.}
\label{fig:rmse}
\end{figure}

\begin{figure}[!tbp]
\centering
  	\includegraphics[width=\textwidth]{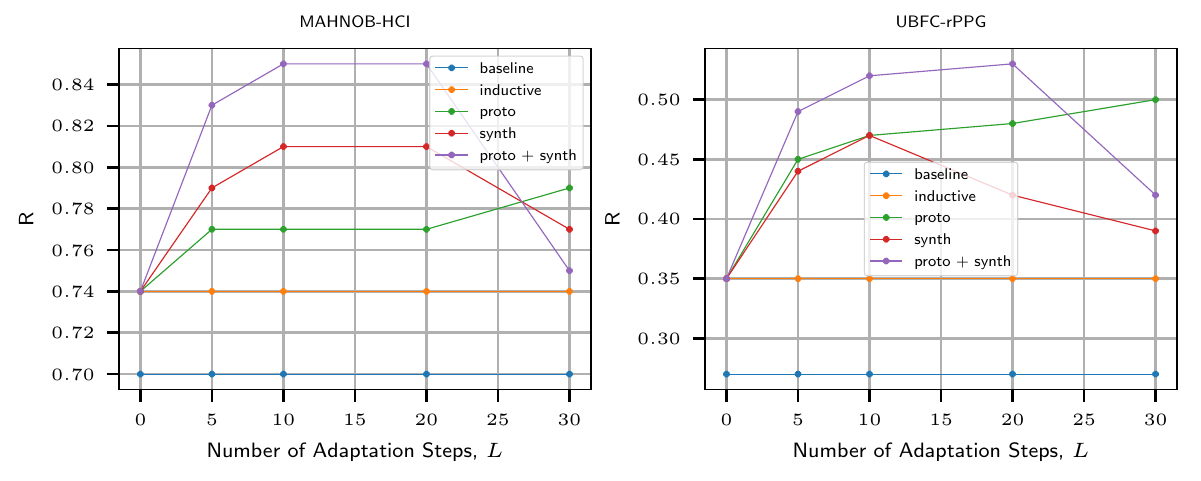}
\caption{Pearson correlation coefficient, R, obtained using different rPPG estimation methods. Demonstrates how the number of adaptation steps, $L$, affects performance.}
\label{fig:r}
\end{figure}

In this section, we study how the number of adaptation steps, $L$, used for transductive inference affects performance. We report results under different metrics, namely, mean absolute error (MAE), standard deviation of error (SD), root mean squared error (RMSE) and Pearson correlation coefficient (R). Tabulated performances for MAHNOB-HCI are shown in Table \ref{tab:mahnob_mae}, \ref{tab:mahnob_sd}, \ref{tab:mahnob_rmse} and \ref{tab:mahnob_r} for MAE, SD, RMSE and R respectively. In the same order, tabulated performances for UBFC-rPPG are shown in Table \ref{tab:ubfc_mae}, \ref{tab:ubfc_sd}, \ref{tab:ubfc_rmse} and \ref{tab:ubfc_r}. Again, using the same order, we show comparison plots in Figure \ref{fig:mae}, \ref{fig:sd}, \ref{fig:rmse} and \ref{fig:r} for both MAHNOB-HCI and UBFC-rPPG.

From the results, we can deduce that the number of adaptation steps used during transductive inference should match the number of steps used during training. This rule only applies to the generation of synthetic gradients for transductive inference but not on the protypical distance minimizer. We hypothesize that the prototypical distance minimizer will eventually converge to some value and doesn't hurt performance if run for infinite number of adaptation steps. This is intuitive as the idea of prototypical distance minimizer is to pull out-of-distribution samples towards the center of the distribution that is modeled by the rPPG estimator using the training data.

\section{Does Joint Adaptation of both Feature Extractor and rPPG Estimator Give Better Results?}
We hypothesize that the estimation of rPPG signal is more efficient if we are able to update the weights of the feature extractor during testing for the adaptation to the new observed distribution. By doing so, we expect that the features generated by the feature extractor fall within the distribution covered by the rPPG estimator, i.e.\ the weights of the rPPG estimator is obtained through the optimization on the training dataset. One might challenge that the joint adaptation of both feature extractor and rPPG estimator weights might result in better performance, contradicting our hypothesis. To demonstrate that our hypothesis holds, we perform an empirical study on whether the joint update approach or the sole update of the weights of the feature extractor performs better. The implementation is straight forward, the synthetic gradient generator is moved towards the output for the joint adaptation case. We show experiments on MAHNOB-HCI using an adaptation steps of 10 in Table \ref{tab:joint}. The empirical results support our hypothesis.

\begin{table}[]
\caption{Results of mean absolute error of HR measurement on MAHNOB-HCI using different adaptation steps, $L$.}
\centering
{\small
\begin{tabular}{lccccc}
\toprule
\multirow{2}{*}{Method} & \multicolumn{5}{c}{MAE of HR (bpm)} \\
\cline{2-6}
                        & $L=0$ & $L=5$ & $L=10$    & $L=20$ & $L=30$ \\
\midrule
End-to-end (baseline)   &   7.47    &   7.47    &   7.47    &   7.47 & 7.47 \\
Meta-rPPG (proto only)  &   7.42    &   6.65    &   6.05    &   6.02 & \textbf{6.02}  \\
Meta-rPPG (synth only)  &   7.42    &   5.00    &   3.88    &   3.70 & 6.25 \\
Meta-rPPG (proto+synth) &   \textbf{7.42}&\textbf{3.89}&\textbf{3.01}&\textbf{3.02}&6.20   \\

 \bottomrule
\end{tabular}
}
\label{tab:mahnob_mae}  
\end{table}

\begin{table}[]
\caption{Results of average standard deviation HR measurement error on MAHNOB-HCI using different adaptation steps, $L$.}
\centering
{\small
\begin{tabular}{lccccc}
\toprule
\multirow{2}{*}{Method} & \multicolumn{5}{c}{SD of HR (bpm)} \\
\cline{2-6}
                        & $L=0$ & $L=5$ & $L=10$    & $L=20$ & $L=30$ \\
\midrule
End-to-end (baseline)   &   \textbf{7.39}    &   7.39    &   7.39    &   7.39 & 7.39 \\
Meta-rPPG (proto only)  &   7.91    &   6.08    &   6.95    &   6.89 & \textbf{6.86}  \\
Meta-rPPG (synth only)  &   7.91    &   5.89    &   5.09    &   4.96 & 7.72 \\
Meta-rPPG (proto+synth) &   7.91&\textbf{4.90}&\textbf{3.68}&\textbf{4.95}&7.81   \\

 \bottomrule
\end{tabular}
}
\label{tab:mahnob_sd}  
\end{table}

\begin{table}[]
\caption{Results of root mean squared error of HR measurement on MAHNOB-HCI using different adaptation steps, $L$.}
\centering
{\small
\begin{tabular}{lccccc}
\toprule
\multirow{2}{*}{Method} & \multicolumn{5}{c}{RMSE of HR (bpm)} \\
\cline{2-6}
                        & $L=0$ & $L=5$ & $L=10$    & $L=20$ & $L=30$ \\
\midrule
End-to-end (baseline)   &   \textbf{8.63}    &   8.63    &   8.63    &   8.63 & 8.63 \\
Meta-rPPG (proto only)  &   8.65    &   6.97    &   6.79    &   6.71 & \textbf{6.67}  \\
Meta-rPPG (synth only)  &   8.65    &   5.15    &   3.96    &   4.02 & 7.11 \\
Meta-rPPG (proto+synth) &   8.65&\textbf{4.65}&\textbf{3.66}&\textbf{3.68}&6.94   \\

 \bottomrule
\end{tabular}
}
\label{tab:mahnob_rmse}  
\end{table}

\begin{table}[]
\caption{Results of Pearson correlation coefficient of HR measurement on MAHNOB-HCI using different adaptation steps, $L$.}
\centering
{\small
\begin{tabular}{lccccc}
\toprule
\multirow{2}{*}{Method} & \multicolumn{5}{c}{R of HR (bpm)} \\
\cline{2-6}
                        & $L=0$ & $L=5$ & $L=10$    & $L=20$ & $L=30$ \\
\midrule
End-to-end (baseline)   &   0.70    &   0.70    &   0.70    &   0.70 & 0.70 \\
Meta-rPPG (proto only)  &   0.74    &   0.77    &   0.77    &   0.77 & \textbf{0.79}  \\
Meta-rPPG (synth only)  &   0.74    &   0.79    &   0.81    &   0.81 & 0.77 \\
Meta-rPPG (proto+synth) &   \textbf{0.74}&\textbf{0.83}&\textbf{0.85}&\textbf{0.85}&0.75   \\

 \bottomrule
\end{tabular}
}
\label{tab:mahnob_r}  
\end{table}

\begin{table}[]
\caption{Results of mean absolute error of HR measurement on UBFC-rPPG using different adaptation steps, $L$.}
\centering
{\small
\begin{tabular}{lccccc}
\toprule
\multirow{2}{*}{Method} & \multicolumn{5}{c}{MAE of HR (bpm)} \\
\cline{2-6}
                        & $L=0$ & $L=5$ & $L=10$    & $L=20$ & $L=30$ \\
\midrule
End-to-end (baseline)   &   \textbf{12.78}    &   12.78    &   12.78    &   12.78 & 12.78 \\
Meta-rPPG (proto only)  &   13.23    &   9.24    &   8.07    &   7.82 & \textbf{7.53}  \\
Meta-rPPG (synth only)  &   13.23    &   11.03    &   7.04    &   9.11 & 11.97 \\
Meta-rPPG (proto+synth) &   13.23&\textbf{7.68}&\textbf{6.07}&\textbf{5.97}&9.82   \\

 \bottomrule
\end{tabular}
}
\label{tab:ubfc_mae}  
\end{table}

\begin{table}[]
\caption{Results of average standard deviation HR measurement error on UBFC-rPPG using different adaptation steps, $L$.}
\centering
{\small
\begin{tabular}{lccccc}
\toprule
\multirow{2}{*}{Method} & \multicolumn{5}{c}{SD of HR (bpm)} \\
\cline{2-6}
                        & $L=0$ & $L=5$ & $L=10$    & $L=20$ & $L=30$ \\
\midrule
End-to-end (baseline)   &   \textbf{13.70}    &   13.70    &   13.70    &   13.70 & 13.70 \\
Meta-rPPG (proto only)  &   14.17    &   10.39    &   9.33    &   9.17 & \textbf{8.23}  \\
Meta-rPPG (synth only)  &   14.17    &   11.00    &   8.37    &   11.92 & 13.94 \\
Meta-rPPG (proto+synth) &   14.17&\textbf{9.01}&\textbf{7.89}&\textbf{7.12}&11.93   \\

 \bottomrule
\end{tabular}
}
\label{tab:ubfc_sd}  
\end{table}

\begin{table}[]
\caption{Results of root mean squared error of HR measurement on UBFC-rPPG using different adaptation steps, $L$.}
\centering
{\small
\begin{tabular}{lccccc}
\toprule
\multirow{2}{*}{Method} & \multicolumn{5}{c}{RMSE of HR (bpm)} \\
\cline{2-6}
                        & $L=0$ & $L=5$ & $L=10$    & $L=20$ & $L=30$ \\
\midrule
End-to-end (baseline)   &   \textbf{13.30}    &   13.30    &   13.30    &   13.30 & 13.30 \\
Meta-rPPG (proto only)  &   14.63    &   10.62    &   9.43    &   9.37 & \textbf{8.90}  \\
Meta-rPPG (synth only)  &   14.63    &   13.41    &   8.55    &   11.55 & 14.62 \\
Meta-rPPG (proto+synth) &   14.63&\textbf{8.92}&\textbf{7.86}&\textbf{7.42}&11.21   \\

 \bottomrule
\end{tabular}
}
\label{tab:ubfc_rmse}  
\end{table}

\begin{table}[]
\caption{Results of Pearson correlation coefficient of HR measurement on UBFC-rPPG using different adaptation steps, $L$.}
\centering
{\small
\begin{tabular}{lccccc}
\toprule
\multirow{2}{*}{Method} & \multicolumn{5}{c}{R of HR (bpm)} \\
\cline{2-6}
                        & $L=0$ & $L=5$ & $L=10$    & $L=20$ & $L=30$ \\
\midrule
End-to-end (baseline)   &   0.27    &   0.27    &   0.27    &   0.27 & 0.27 \\
Meta-rPPG (proto only)  &   0.35    &   0.45    &   0.47    &   0.48 & \textbf{0.50}  \\
Meta-rPPG (synth only)  &   0.35    &   0.44    &   0.47    &   0.42 & 0.39 \\
Meta-rPPG (proto+synth) &   \textbf{0.35}&\textbf{0.49}&\textbf{0.52}&\textbf{0.53}&0.42   \\

 \bottomrule
\end{tabular}
}
\label{tab:ubfc_r}  
\end{table}

\begin{table}[]
\caption{Results of average HR measurement on MAHNOB-HCI comparing the difference between joint adaptation of \textit{both feature extractor and rPPG estimator} and updating the \textit{feature extractor only}.}
\centering
{\small
\begin{tabular}{lccc}
\toprule
\multirow{2}{*}{Method} & \multicolumn{3}{c}{HR (bpm)} \\
\cline{2-4}
                                                & MAE       & RMSE      & R \\
\midrule
Joint Adaptation          &   4.25       &   6.09    &   0.81 \\
Extractor Only          &   \textbf{3.01}       &   \textbf{3.68}    &   \textbf{0.85} \\
 \bottomrule
\end{tabular}
}
\label{tab:joint}  
\end{table}

\section{Visualization of Feature Activation Map Using Different Methods}
In this section, we show the visualization for feature activation map using the methods we introduced and is compared with the baseline that uses an end-to-end supervised learning method. Ablation study is also performed by showing feature activation maps obtained using individual methods that we proposed. More subjects are also shown here to give a better understanding of the importance of transductive inference for rPPG estimation using a deep learning model.

\section{Demonstration Using Video}
We show the implementation of our algorithm on videos extracted from MAHNOB-HCI to show its performance during deployment. We demonstrate our algorithm on videos of 3 subjects and a snapshot of a single frame from one of the video is shown in Fig. \ref{fig:video}. From the video attached in the supplementary materials, it can be observed that the feature activation maps corresponding to our transductive inference method is relatively consistent when compared to a model trained in an end-to-end fashion. Please refer to our video for a better understanding on the improvements brought by the introduction of transductive inference to rPPG estimation.

\begin{figure}[!tbp]
\centering
  	\includegraphics[width=\textwidth]{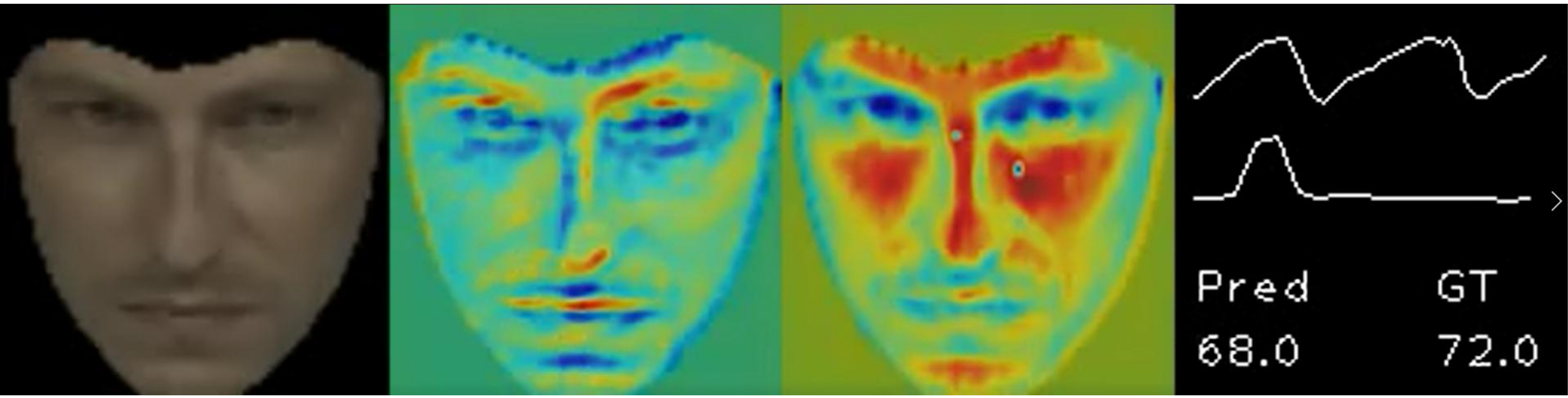}
\caption{A frame extracted from a video from MAHNOB-HCI. From left to right: 1. Pre-processed face image (zero-ing of pixels outside facial landmarks), 2. feature activation maps corresponding to end-to-end trained model, 3. feature activation map corresponding to Meta-rPPG (proto+synth) and 4. plots containing rPPG signal (top), power spectral density of rPPG signal (middle), predicted (bottom-left) and ground truth heart rate (bottom-right) in beats-per-minute (BPM).}
\label{fig:video}
\end{figure}

\begin{figure}[!tbp]
\centering
  	\includegraphics[width=\textwidth]{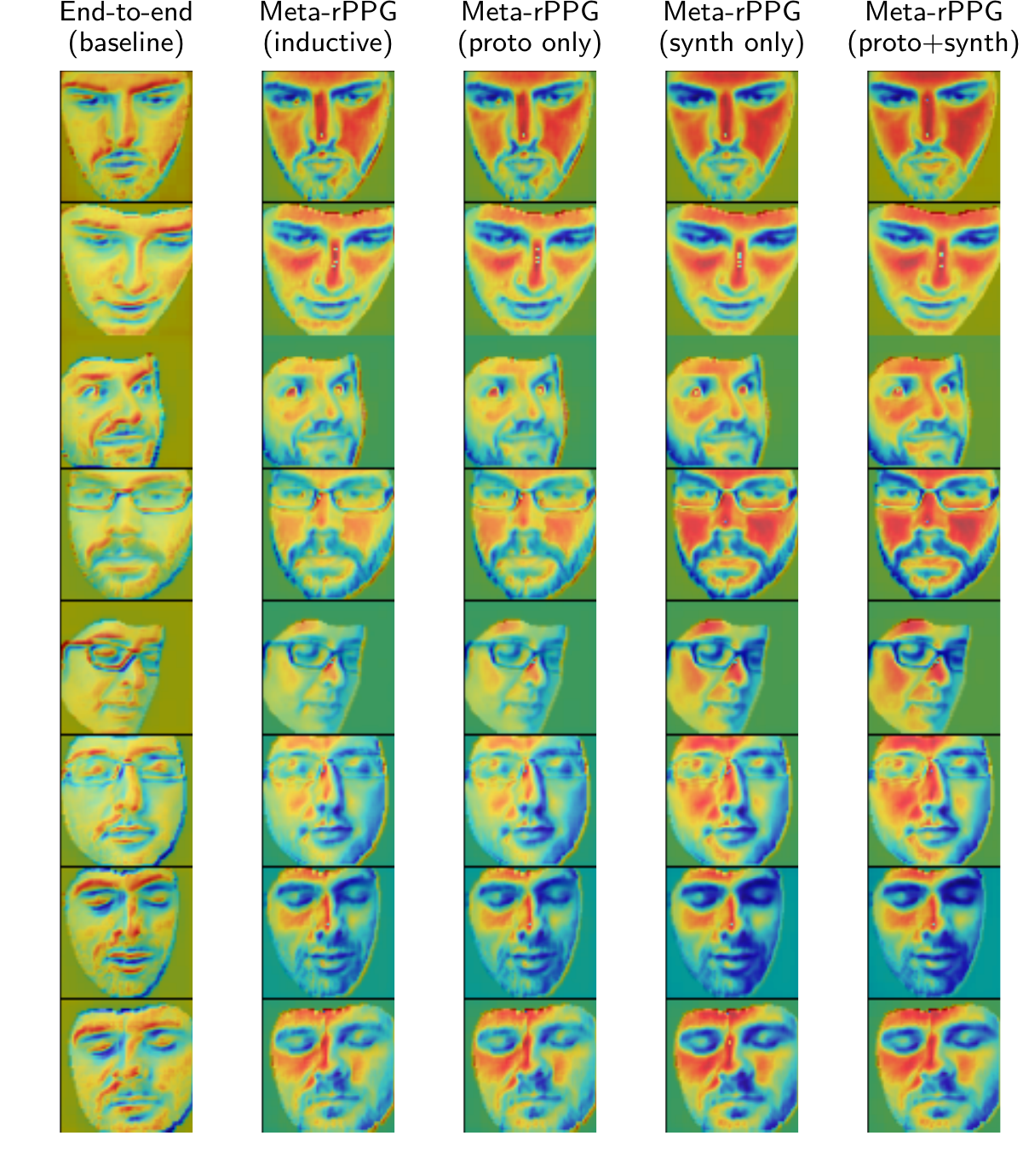}
\caption{Feature activation map visualization of 8 subjects (each row corresponds to one subject) using different training methods. Ablation study is performed by inspecting the feature map activation the results upon the application of every proposed transductive inference method. Usage of transductive inference results in activations of higher contrast and covers larger region of facial features that contributes to rPPG estimation.}
\label{fig:fvm}
\end{figure}

\end{document}